\documentclass[letterpaper, 10 pt, conference]{ieeeconf}  

\IEEEoverridecommandlockouts                              

\usepackage{graphics} 
\usepackage{epsfig} 
\usepackage{amsmath} 
\usepackage{amssymb}  
\usepackage[table]{xcolor}

\title{\LARGE \bf
SCAL for Pinch-Lifting: Complementary Rotational and Linear Prototypes for Environment-Adaptive Grasping*
}

\author{Wentao Guo and Wenzeng Zhang, \textit{Member, IEEE}
\thanks{*Research supported by Foundation of \textit{Enhanced Student Research Training (E-SRT)} and Foundation of \textit{Open Research for Innovation Challenges (ORIC)}, X-Institute.}
\thanks{Wentao Guo is with Computer Science and Technology, Beijing Institute of Technology, China and Laboratory of Robotics, X-Institute, Shenzhen, China. (email: yinsumirage@gmail.com)}
\thanks{Wenzeng Zhang is with Laboratory of Robotics, X-Institute, Shenzhen, China and Dept. of Mechanical Engineering, Tsinghua University, Beijing, China (Corresponding author, email: wenzeng75@163.com).}
}

\begin{document}

\maketitle
\thispagestyle{empty}
\pagestyle{empty}

\begin{abstract}

This paper presents environment-adaptive pinch-lifting built on a slot-constrained adaptive linkage (SCAL) and instantiated in two complementary fingers: SCAL-R, a rotational-drive design with an active fingertip that folds inward after contact to form an envelope, and SCAL-L, a linear-drive design that passively opens on contact to span wide or weak-feature objects. Both fingers convert surface following into an upward lifting branch while maintaining fingertip orientation, enabling thin or low-profile targets to be raised from supports with minimal sensing and control. Two-finger grippers are fabricated via PLA-based 3D printing. Experiments evaluate (i) contact-preserving sliding and pinch-lifting on tabletops, (ii) ramp negotiation followed by lift, and (iii) handling of bulky objects via active enveloping (SCAL-R) or contact-triggered passive opening (SCAL-L). Across dozens of trials on small parts, boxes, jars, and tape rolls, both designs achieve consistent grasps with limited tuning. A quasi-static analysis provides closed-form fingertip-force models for linear parallel pinching and two-point enveloping, offering geometry-aware guidance for design and operation. Overall, the results indicate complementary operating regimes and a practical path to robust, environment-adaptive grasping with simple actuation.

\end{abstract}

\section{INTRODUCTION}

In unstructured environments, grippers often operate in proximity to environmental constraints such as tabletops, ramps, steps, or walls. Rather than avoiding these contacts, leveraging them can render thin, low-profile, or surface-adhered objects graspable through sliding, scooping, or ramp-ascent motions. Such environment-assisted strategies can expand the feasible grasp set while reducing perception and planning demands~\cite{exploit2015}.

To this end, the field has advanced along two main lines. One seeks multi-finger, high-DOF dexterous hands. The other develops underactuated and compliant mechanisms that embed passive “mechanical intelligence” to reduce the need for high-bandwidth sensing and control~\cite{underactuated2007}. Underactuation—fewer actuators than kinematic DOFs—typically relies on springs, compliant joints, and mechanical end-stops to provide constraints, striking a balance among control burden, operational capability, cost and weight~\cite{AnalysisUnderactuated2011}. Representative underactuated grippers from industry and academia (e.g., Robotiq~\cite{robotiq}, GR2~\cite{GR2}, uGRIPP~\cite{uGRIPP}, ParaGripper~\cite{ParaGripper}) have been extensively studied for parallel pinching, grasping, and in-hand manipulation, with much of this literature assuming “floating” targets or designing grasps that are parallel to the tabletop~\cite{PASA2016}, leaving the systematic modeling and exploitation of environmental contact comparatively underexplored.

When environmental interactions are considered explicitly in gripper design, recent approaches roughly fall into three categories:

(1) \textit{Perception/control}: These methods use force, proximity, or vision sensing with impedance or force control to detect contact and plan surface-following insertion, offering high generality but requiring precise calibration and high control bandwidth. Examples include proximity-aware tabletop grasping~\cite{proximity2018} and recent proximity-tactile few-shot tool-use policies~\cite{tool2025}.

(2) \textit{Soft/high-compliance}: Materials with inherent deformability or flexible transmissions conform to surface irregularities and tolerate pose errors, naturally enabling smooth contact transitions but limiting stiffness and repeatability. For example, the tendon-driven Velo gripper can pinch thin objects despite fingertip-table contact~\cite{Velo2014}, and the belt-driven BLT gripper distributes contact forces and actively transitions from pinch to grasp~\cite{BLT2020}.

(3) \textit{Linkage-driven underactuated adaptation (our focus)}: Linkages, slots, springs, and stops embed passive “mechanical intelligence” to achieve contact-preserving mode switching and displacement compensation, enabling along-surface sliding, alignment, and surface-to-lift transitions without sensing-heavy control. The remainder of this paper focuses on this direction.

A key operation in this line is \textit{pinch-lifting}, which first establishes stable contact on a support surface and then lifts the object; unlike simple vertical lifting, it couples a near-horizontal sliding phase with an upward phase and thus leverages environmental contacts. Yoon et al. showed that a four-bar fingertip slider can steer the net force from tangential to vertical, outlining sliding/ascending/lifting regimes and design trade-offs~\cite{analysis2021}. Earlier, a three-finger adaptive hand combined this pinch-lifting behavior with an active fingertip module that folds inward after contact, enabling environment-adaptive grasping and active pinch-to-envelope transitions~\cite{three-finger2021}. 

Complementing these rotational designs, a fully passive finger with horizontal drive and vertical compliance achieves press-down and scooping on supports~\cite{omega2022}. Although it does not perform surface-to-lift transitions, it highlights a linear-drive paradigm and a compliance dimension that could be combined with pinch-lifting. Collectively, these studies establish theoretical and empirical bases for exploiting environmental contacts—including pinch-lifting—yet a generalizable design framework spanning rotational and linear drives remains lacking.

\begin{figure}[t]
    \centering
    \includegraphics[width=0.98\linewidth]{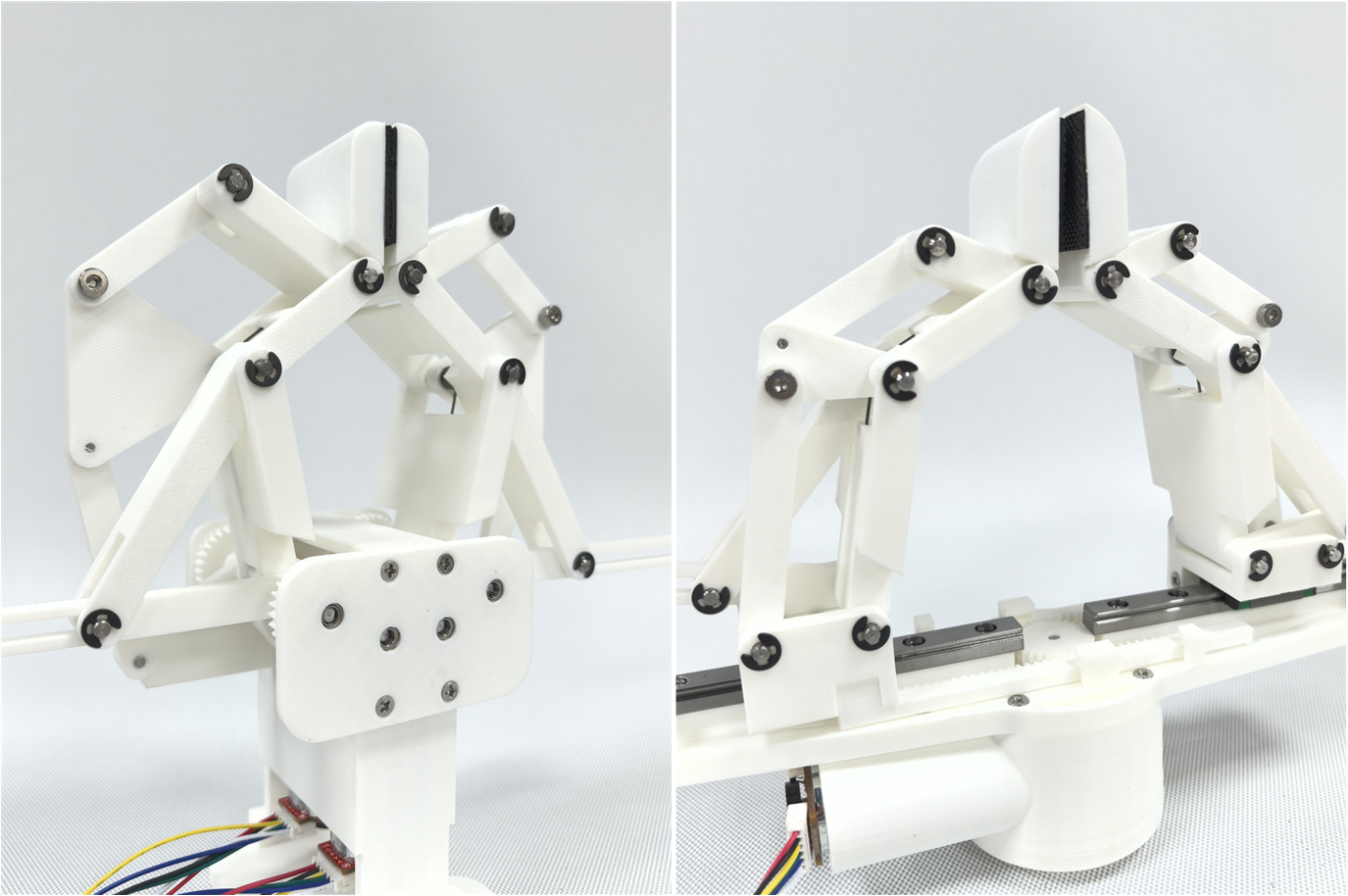}
    \caption{3D-printed prototype of the proposed SCAL-R and SCAL-L.}
    \label{fig:prototype}
\end{figure}

Motivated by pinch–lift, we instantiate a Slot-Constrained Adaptive Linkage (SCAL)—a slotted-guide linkage that shapes an oblique upward–outward fingertip path—as a common kinematic primitive across two actuation regimes, yielding two complementary prototypes (see Fig.~\ref{fig:prototype}):

\begin{enumerate}
    \item \textbf{Rotational drive (SCAL-R): environment-adaptive with an active fingertip.}
    SCAL kinematics generate an upward–outward fingertip trace; a base-in/tip-out compensation mitigates lateral pushing on the support surface so the net contact wrench rotates toward lift. After lift-off, a compact active fingertip module folds outward to form an envelope, improving post-lift stability and enlarging the capture range; the mechanism negotiates ramps and steps effectively.

    \item \textbf{Linear drive (SCAL-L): exploratory contact in a parallel frame.}
    Two vertical fingertips are mounted on a linearly driven parallel frame. Reusing the same SCAL, the fingertip follows an upward–outward trace that realizes a probe–slide–lift transition under surface contact. A contact-triggered passive opening widens an initially small aperture to conform to the object surface before closing, preserving control simplicity while accommodating thin edges and surface-adhered targets.
\end{enumerate}

These SCAL-based prototypes let us isolate how actuation mode and contact strategy govern robust surface-to-lift transitions in practice and provide the physical platforms previewed in Fig.~\ref{fig:prototype} for the modeling and quantitative evaluation that follow.

\section{DESIGN CONCEPT}
\label{sec:design_concept}

In this section we introduce a Slot-Constrained Adaptive Linkage (SCAL)—a linkage whose coupler is constrained by a slider pin in a machined slot—to generate an upward-outward fingertip path that preserves surface contact and enables robust surface-to-lift transitions. We then instantiate this template in two complementary prototypes: SCAL-R (rotational drive with an active fingertip that envelopes after lift-off) and SCAL-L (linear drive with a contact-triggered passive opening for thin edges and large-rim objects).

\subsection{Slot-Constrained Adaptive Linkage (SCAL)}

The proposed \emph{slot-constrained adaptive linkage (SCAL)} consists of a ground pivot at $A=(0,0)$, an input link $AC$ constrained by a straight slot (rotated about $A$ in the rotational prototype and translated along the slot in the linear prototype), a follower link $AB$ closed by kinematics, and a bent distal link $CBD$ that carries the fingertip at $D$ with a fixed interior angle $\gamma$ at $B$. This bent link can be regarded as a single continuous member deflected at $B$ rather than two separate links. 

\begin {figure}[t]
    \centering
    \includegraphics [width=0.8\linewidth]{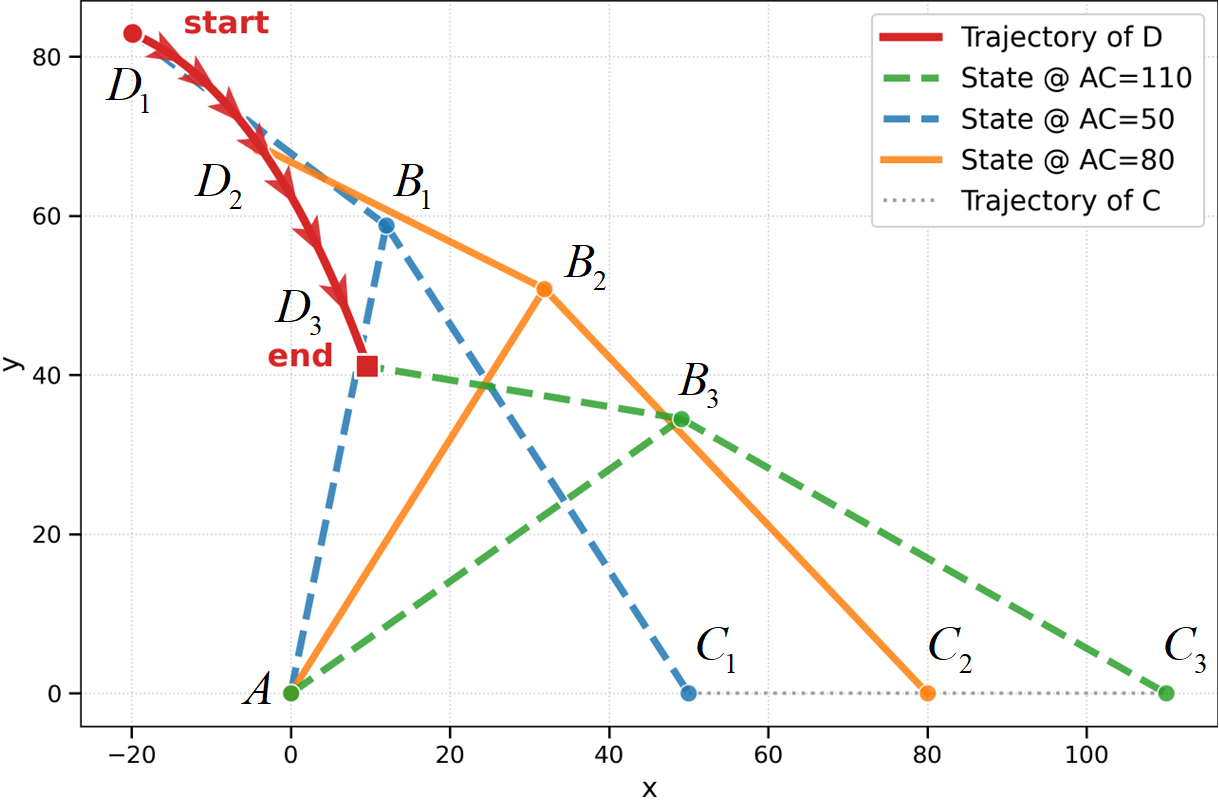}
    \caption {Motion Analysis of the SCAL.}
    \label {fig:scal_motion}
\end {figure}

Figure~\ref{fig:scal_motion} shows the simulated motion of the SCAL. Geometry is specified by normalized ratios using a base length $l$, as listed in Table~\ref{tab:scal-params}. The mechanism starts from the \emph{minimum} slot length (i.e., $C$ at the near end), and $|AC|$ then increases as the input. As $|AC|$ varies from $5l$ to $11l$, loop-closure uniquely determines $B$ and $D$; the locus of $D$ is plotted in red. In this schematic the trace appears \emph{downward-and-outward}; in tabletop deployment the base $A$ is mounted above while $D$ contacts the object below, so the same trace corresponds to an \emph{upward-and-outward} approach relative to the surface, which is exploited for contact preservation and the subsequent surface-to-lift transition.

\begin{table}[t]
    \centering
    \caption{Normalized geometric parameters of the SCAL (base unit $l$).}
    \label{tab:scal-params}
    \rowcolors{1}{blue!10}{yellow!15}
    \begin{tabular}{|c|c|c|c|c|}
        \hline
        $|AB|$ & $|CB|$ & $|BD|$ & $\gamma$ & $|AC|$ range \\ \hline
        $6\,l$ & $7\,l$ & $4\,l$ & $160^\circ$ & $5\,l\!-\!11\,l$ \\ \hline
    \end{tabular}
\end{table}

\subsection{SCAL Finger Architecture}

We embed the SCAL core ($A$–$B$–$C$–$D$) into a finger-level assembly that preserves fingertip orientation and realizes passive adaptation after contact. In phalange terminology, $AB$ acts as the \emph{proximal phalange}, $BD$ as the \emph{intermediate phalange}, and the phalange $DI$ as the \emph{distal phalange}. The SCAL subassembly provides the passive degree of freedom: once the fingertip at $D$ touches the object, the input link $AC$ moves away from the slot’s near end and \emph{opens} along the slot, which—through the bent distal member $CBD$—causes $D$ to \emph{fold outward}. Because the SCAL path is oblique, the same motion produces an upward–outward branch under contact, preparing the subsequent surface-to-lift transition.

\begin{figure}[t]
  \centering
  \includegraphics[width=0.7\linewidth]{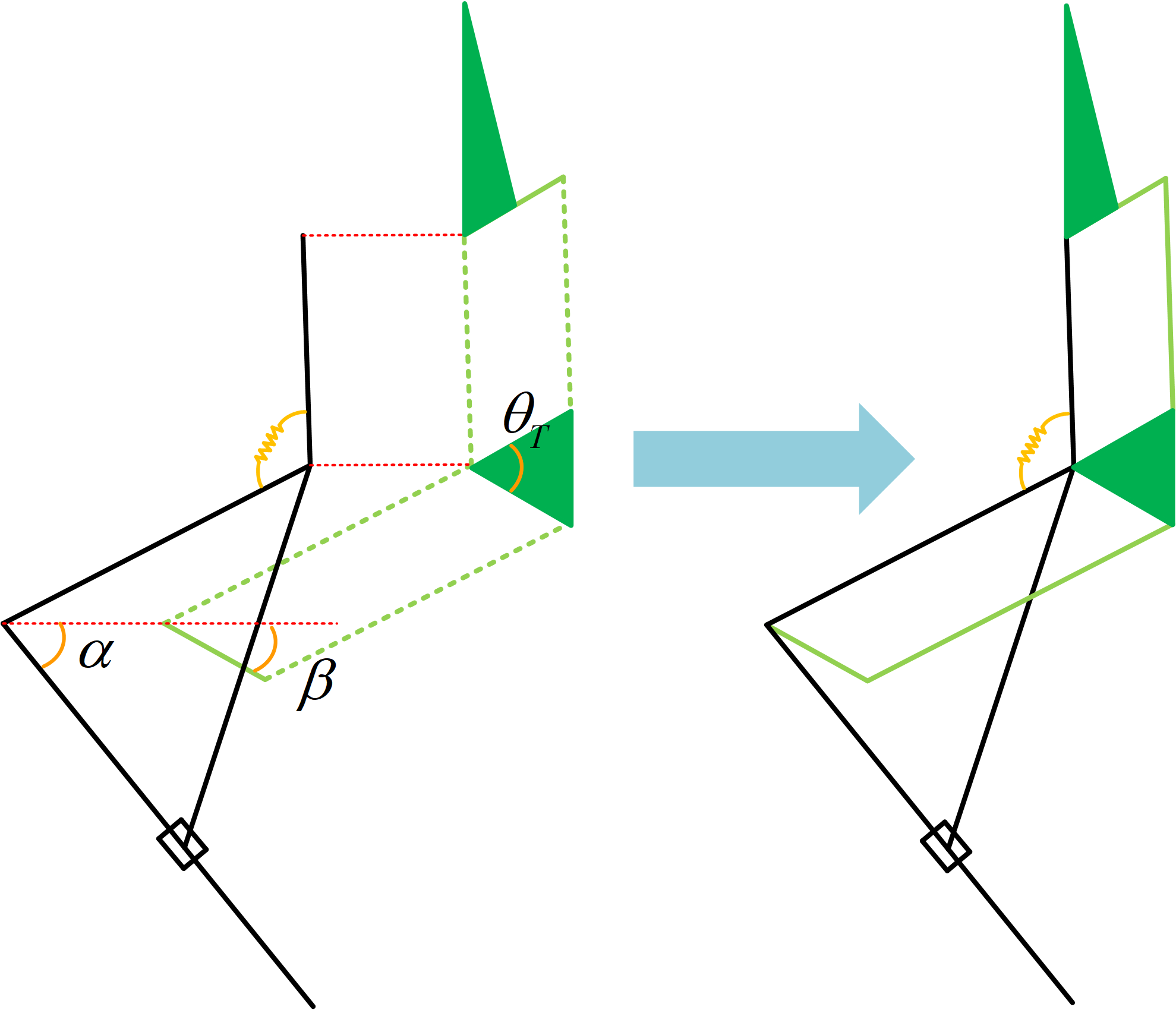}
  \caption{SCAL finger.}
  \label{fig:scal_finger_arch_combo}
\end{figure}

To maintain a fixed fingertip orientation while the SCAL executes its path, we surround the core with two parallelogram frames and a rigid triangular connector. A lower parallelogram $ABFE$ anchors the proximal frame at $A$ and carries a short side $AE$; an upper parallelogram $BDHG$ anchors the intermediate frame at $B$ and carries a short side $DH$. A triangular connector $BFG$ couples the two frames. With the chosen geometry—both parallelograms sharing the same short-side length and the triangular connector set as an isosceles element—the orientation imposed at $AE$ is transmitted to $DH$ in a near one-to-one fashion. Since $DI$ is rigidly fixed to $DH$, fixing the bearing of $AE$ effectively fixes the distal phalange; conversely, actuating $AE$ provides an immediate route to \emph{active} distal orientation control when desired. As sketched in Fig.~\ref{fig:scal_finger_arch_combo}, three preset angles $\alpha$, $\beta$, and $\theta_t$ determine the initial setup; their values will be specified later for the two prototypes (SCAL-R / SCAL-L).

A torsion spring biases the input link toward the slot’s near end so that, in the default state, $C$ stays at the minimum $|AC|$. When the fingertip at $D$ contacts an object, reaction forces drive $AC$ outward against the spring preload, allowing the fingertip to conform while maintaining contact. The complete finger is later mirrored to form a two-finger gripper in both prototypes.

\begin{table}[t]
  \centering
  \caption{Finger-level geometric parameters (base unit $l$). Core SCAL ratios are listed in Table~\ref{tab:scal-params}.}
  \label{tab:scal_finger_params}
  \rowcolors{1}{blue!10}{yellow!15}
  \begin{tabular}{|c|c|c|c|}
    \hline
    $|AE|$ & $|BF|$ & $|BG|$ & $|DI|$ \\ \hline
    $2\,l$ & $2\,l$ & $2\,l$ & $4\,l$ \\ \hline
  \end{tabular}
\end{table}

\subsection{SCAL-R: Rotational Drive with Active Fingertip}

Based on the finger architecture described above, the SCAL-R prototype instantiates the mechanism with a base unit of $l=10\,\mathrm{mm}$. The initial bearing of the lower parallelogram short side $AE$ is set to $\alpha=30^\circ$ from the horizontal, the input link $AC$ is initially at $\beta=78.46^\circ$, and the triangular connector is configured as an isosceles triangle with an apex angle of $\theta_T=60^\circ$. These values determine the phase relation between the SCAL core, the parallelogram frames, and the fingertip orientation; their counterparts in the SCAL-L prototype will be described later. 

\begin{figure}[t]
    \centering
    \includegraphics[width=0.7\linewidth]{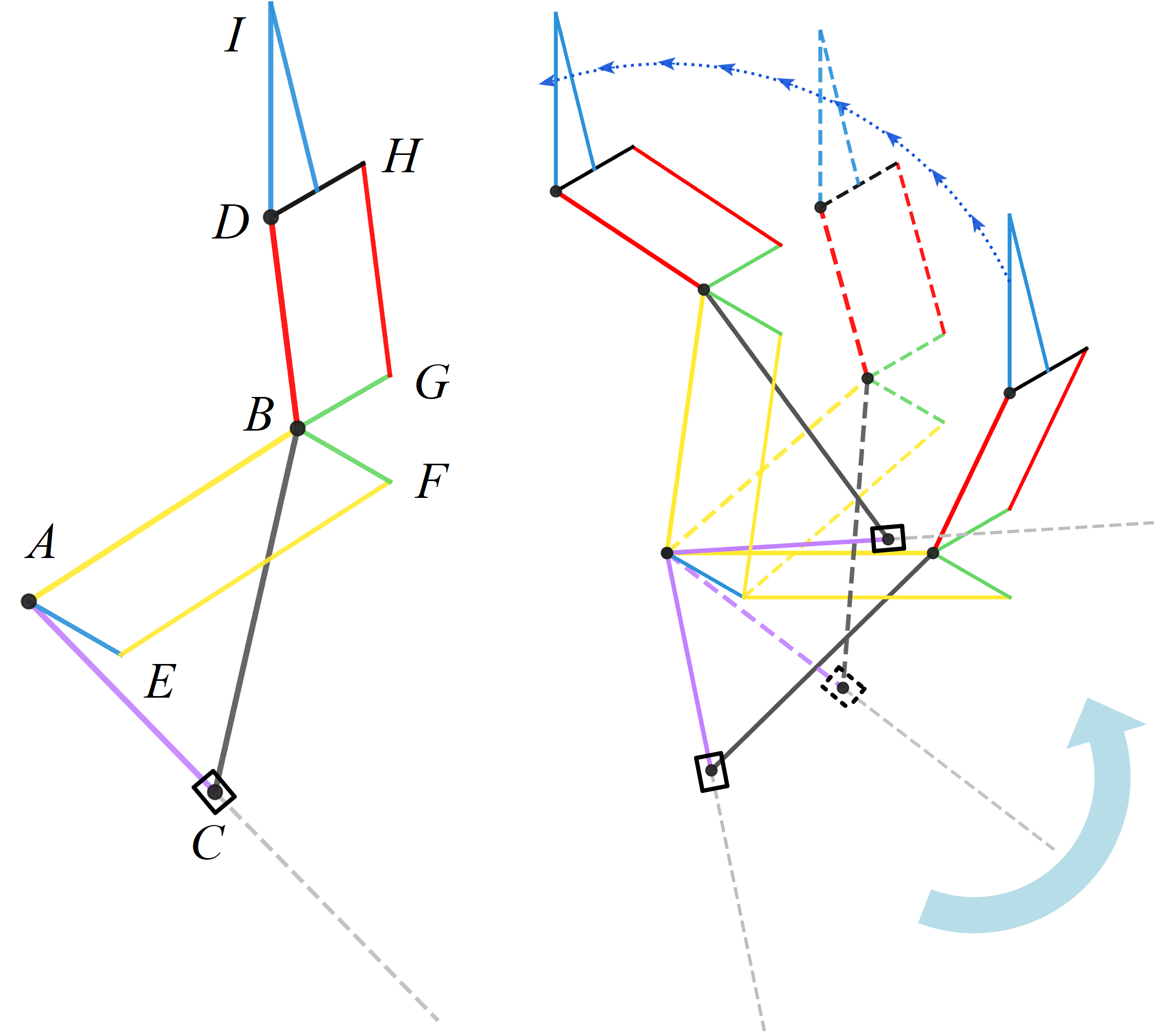}
    \caption{SCAL-R and its free-space motion.}
    \label{fig:rotato}
\end{figure}

Figure~\ref{fig:rotato} shows the free-space motion of SCAL-R. The left panel depicts a simplified configuration with annotations; the right panel plots the rotation path. With the torsion spring holding $C$ at the slot’s near end, the linkage behaves as a rigid assembly during the initial approach: the SCAL core ($A$–$B$–$C$–$D$), the parallelogram frames, and the distal phalange $DI$ rotate together about the base pivot $A$ while the fixed bearing of $AE$ keeps the fingertip vertical. 

\begin{figure}[t]
    \centering
    \includegraphics[width=0.95\linewidth]{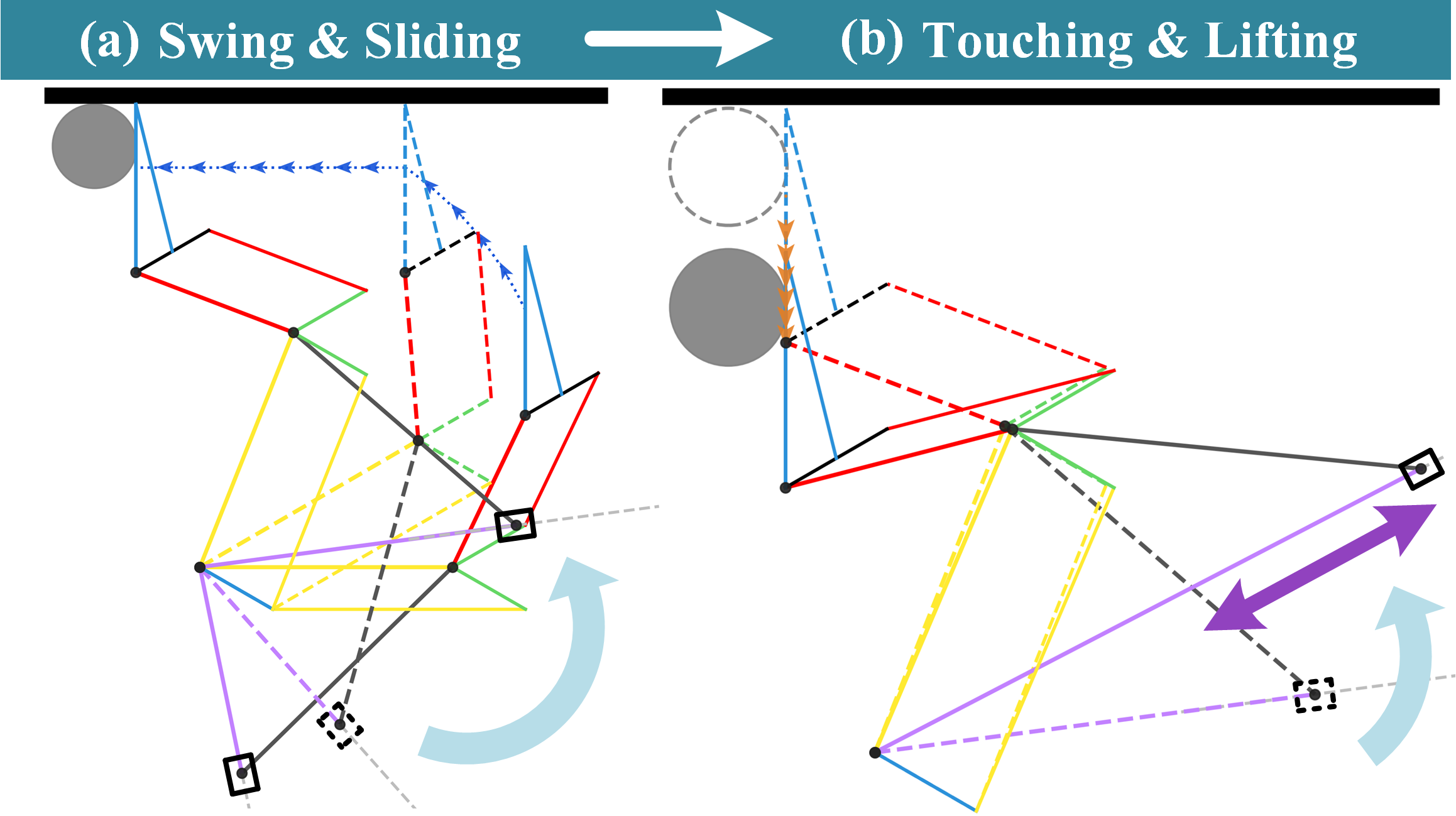}
    \caption{SCAL-R environment-assisted pinch-lifting: (a) swing \& sliding, (b) touching \& lifting.}
    \label{fig:rotato_env}
\end{figure}

When the fingertip at $I$ contacts the tabletop (Fig.~\ref{fig:rotato_env}), the reaction pushes $C$ outward along the slot against the spring preload, triggering passive adaptation: (a) swing \& sliding along the surface with parallel pinching; (b) touching \& lifting as the path bends upward. This two-phase sequence enables thin, low-profile objects to be scooped and lifted by leveraging contact.

\begin{figure}[t]
    \centering
    \includegraphics[width=0.8\linewidth]{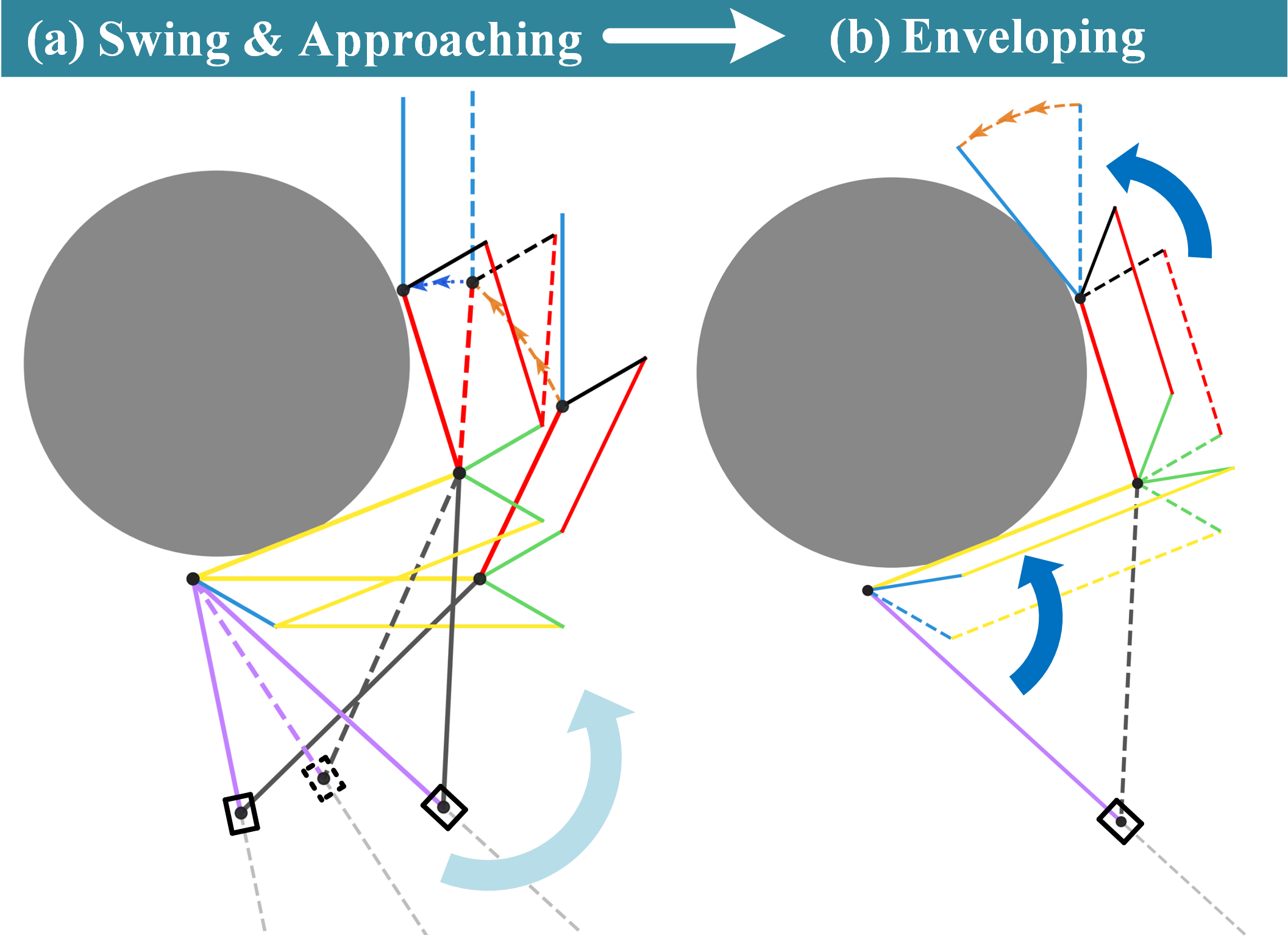}
    \caption{SCAL-R enveloping grasp: (a) swing \& approaching, (b) enveloping.}
    \label{fig:rotato_wrap}
\end{figure}

The second behavior arises with bulky objects (Fig.~\ref{fig:rotato_wrap}). The proximal linkage swings to approach and initiate contact, after which the intermediate phalange establishes a stabilizing support while the slot link continues advancing the fingertip toward the object. Once a firm two-point contact is formed, the active fingertip executes a commanded fold, transitioning the grasp from pinch to envelope; this closure increases the normal force and contact wrench, securing the object for lift-off and subsequent transport.

\subsection{SCAL-L: Linear Drive with Passive Opening}

The SCAL-L prototype reuses the same finger architecture but mounts the SCAL linkage on a horizontally sliding carriage. The lower frame $AE$ and the input link $AC$ are rigidly fixed to this carriage and translate as a whole, so the base pivot $A$ moves with the carriage instead of being anchored. In the initial configuration, $AE$ lies horizontal ($\alpha=0^\circ$), $AC$ is tilted slightly upward at $\beta=-15^\circ$, and the triangular connector is set to an apex angle of $\theta_T=20^\circ$. A torsion spring biases $C$ toward the near end of the slot, making the linkage behave as a rigid assembly during free-space translation.

\begin{figure}[t]
    \centering
    \includegraphics[width=0.85\linewidth]{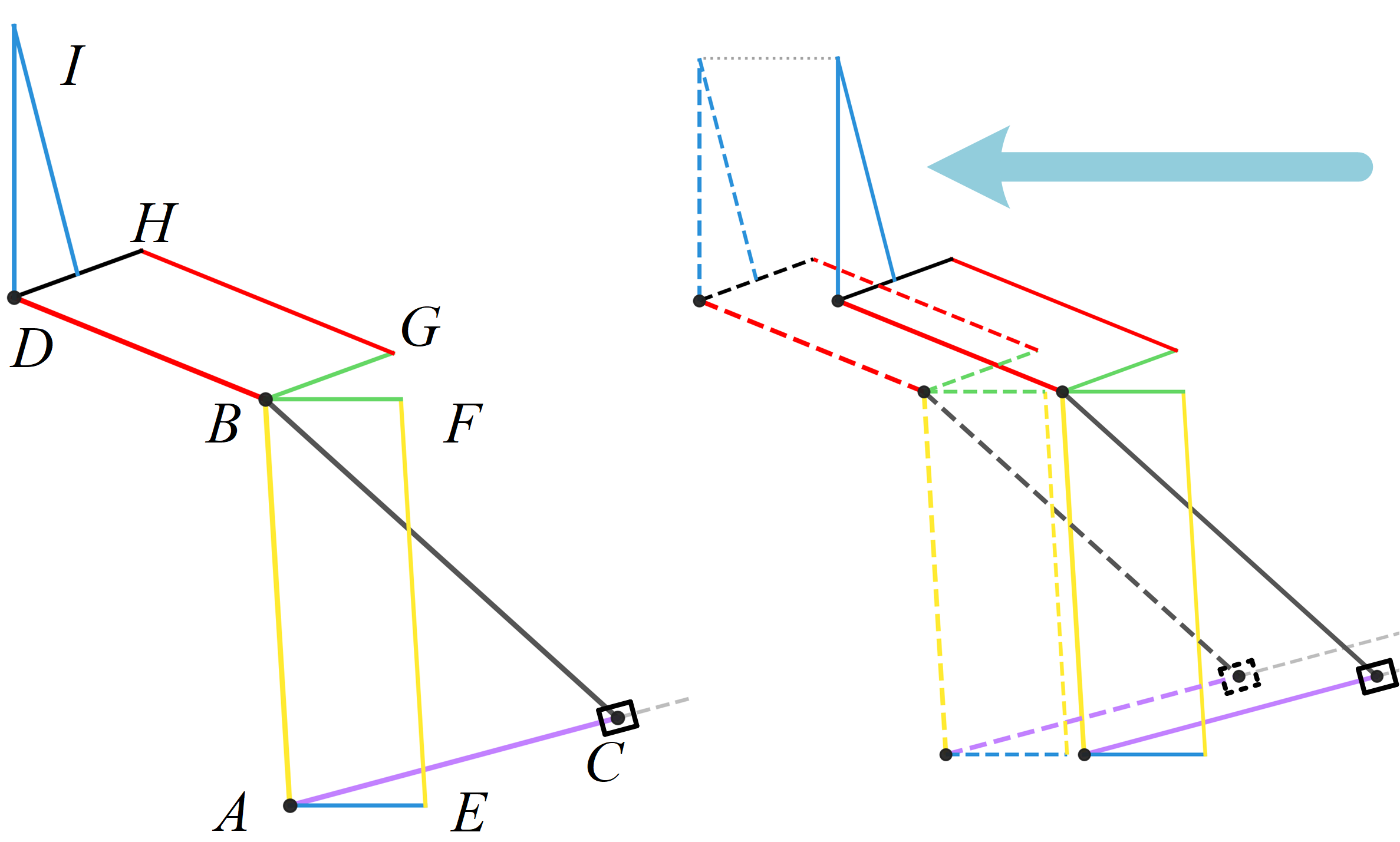}
    \caption{SCAL-L and its free-space motion.}
    \label{fig:lin_overview}
\end{figure}

Figure~\ref{fig:lin_overview} shows the assembled SCAL-L finger and its basic horizontal actuation. The left panel is a simplified schematic with key members; the right panel plots the horizontal sweep path. In free space, the whole finger translates while the distal phalange remains vertical due to the parallelogram-triangular-connector chain. A torsion spring biases the slot slider inward, so the assembly acts rigid during approach.

\begin{figure}[t]
    \centering
    \includegraphics[width=0.95\linewidth]{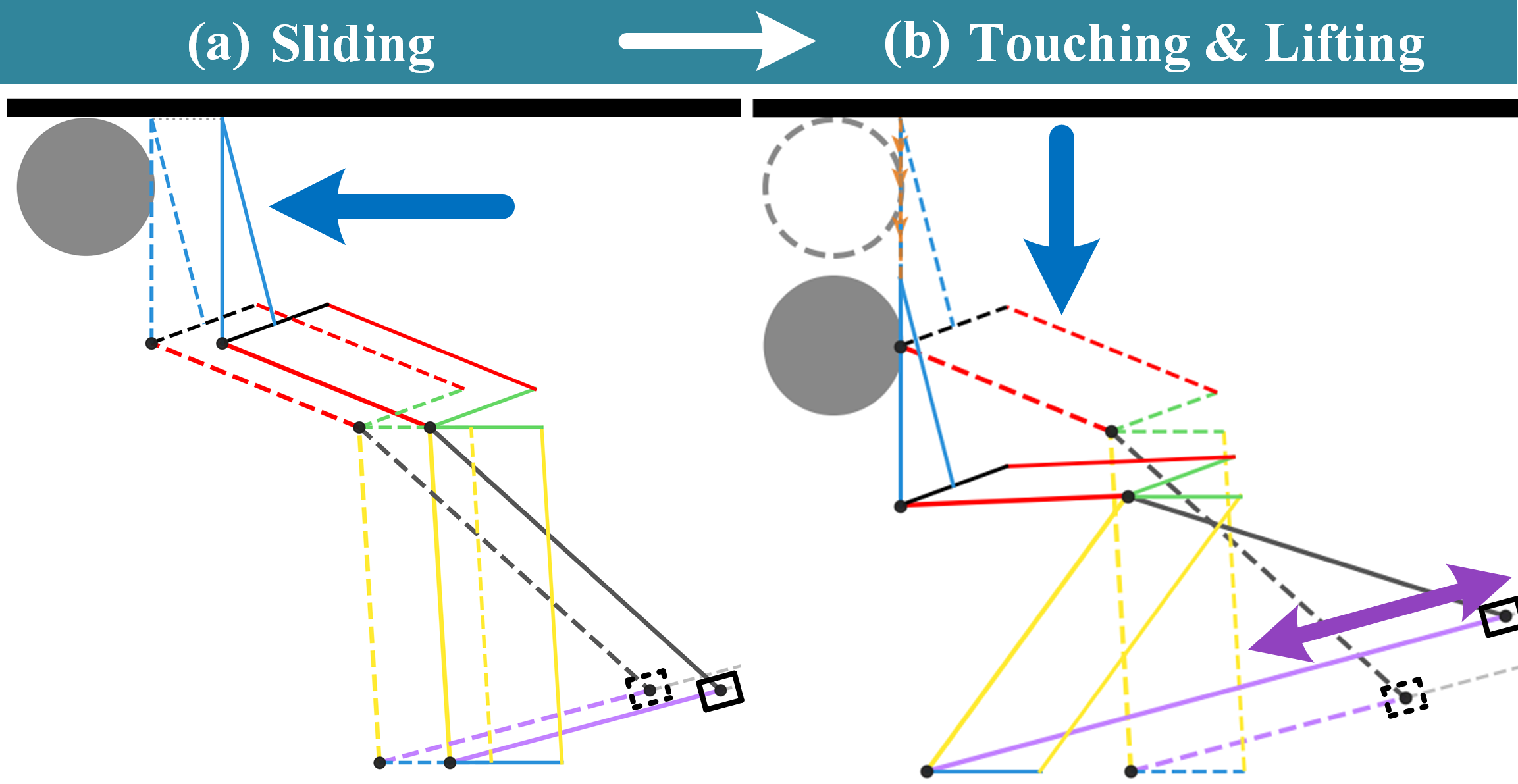}
    \caption{SCAL-L environment-assisted pinch-lifting: (a) sliding, (b) touching \& lifting.}
    \label{fig:lin_lift}
\end{figure}

When the fingertip $I$ encounters a target along the horizontal sweep, the contact reaction drives $C$ outward along the slot against the spring preload. The SCAL then passively adapts while maintaining contact, and as the carriage continues to translate, the fingertip path bends upward, lifting the object off the support (Fig.~\ref{fig:lin_lift}). This is the same environment-assisted pinch-lifting mechanism as in SCAL-R, realized here with translation rather than rotation.

\begin{figure}[t]
    \centering
    \includegraphics[width=0.95\linewidth]{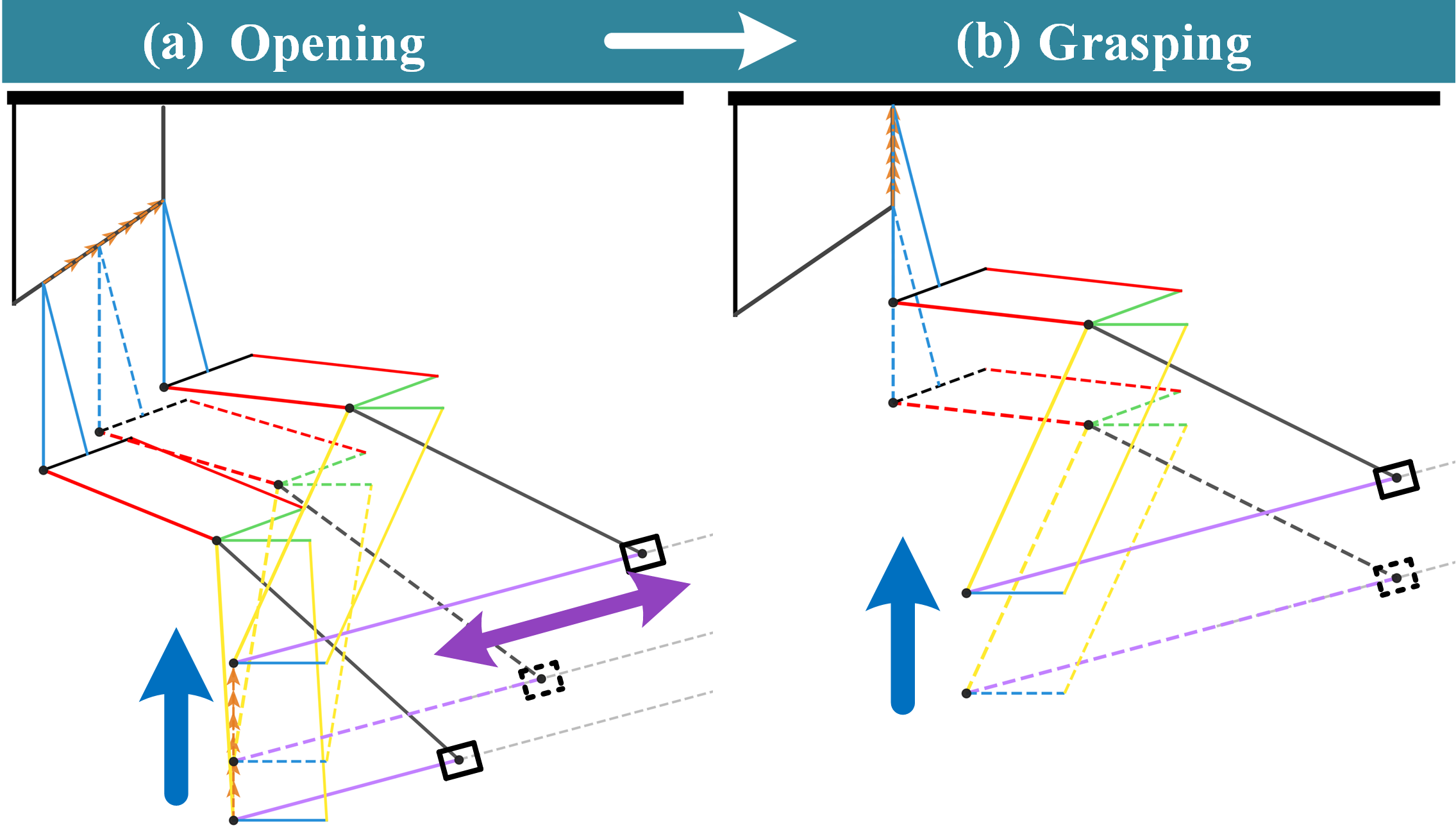}
    \caption{SCAL-L exploratory-contact grasping: (a) opening, (b) grasping.}
    \label{fig:lin_explore}
\end{figure}

In addition, the SCAL-L finger can grasp objects wider than its initial aperture via contact-triggered passive opening (Fig.~\ref{fig:lin_explore}). With the horizontal drive idle, the finger is lowered to probe; upon contact, the slot slider is pushed outward to widen the aperture, after which the horizontal drive closes to secure the grasp. This exploratory-contact mode suits irregular shapes and large rims.

\section{MECHANICAL IMPLEMENTATION}

This section instantiates the proposed designs in SolidWorks. Building on the normalized geometry in Sec.~\ref{sec:design_concept}, we set the base unit to $l=10\,\mathrm{mm}$ and realize two finger prototypes that share all link lengths and slot travel; the only geometric differences are the initial bearings and the triangular connector apex. Table~\ref{tab:mech-params} summarizes the physical dimensions and the prototype-specific angles used in our builds.

\begin{table}[t]
  \centering
  \caption{Consolidated mechanical parameters (base unit $l=10\,\mathrm{mm}$).}
  \label{tab:mech-params}
  \renewcommand{\arraystretch}{1.1}
  \begin{tabular}{|l|c|}
    \hline
    \rowcolor{blue!10} 
    \textbf{Parameter} & \textbf{Value} \\\hline
    \rowcolor{yellow!20}
    $|AB|,\,|CB|,\,|BD|$ [mm] & $60,\,70,\,40$ \\\hline
    \rowcolor{yellow!15}
    $|AC|$ range [mm] & $[50,\,110]$ \\\hline
    \rowcolor{yellow!15}
    $|AE|,\,|BF|,\,|BG|$ [mm] & $20,\,20,\,20$ \\\hline
    \rowcolor{yellow!15}
    $|DI|$ [mm] & $40$ \\\hline
    \rowcolor{blue!10} 
    \multicolumn{2}{|l|}{\textbf{Prototype-specific initial angles} [deg]} \\\hline
    \rowcolor{yellow!15}
    SCAL-R $(\alpha,\beta,\theta_T)$ & $(30,\,78.463,\,60)$ \\\hline
    \rowcolor{yellow!15}
    SCAL-L $(\alpha,\beta,\theta_T)$ & $(0,\,-15,\,20)$ \\\hline
  \end{tabular}
\end{table}

\subsection{SCAL-R: Mechanical Design}

\begin{figure}[t]
  \centering
  \includegraphics[width=0.95\linewidth]{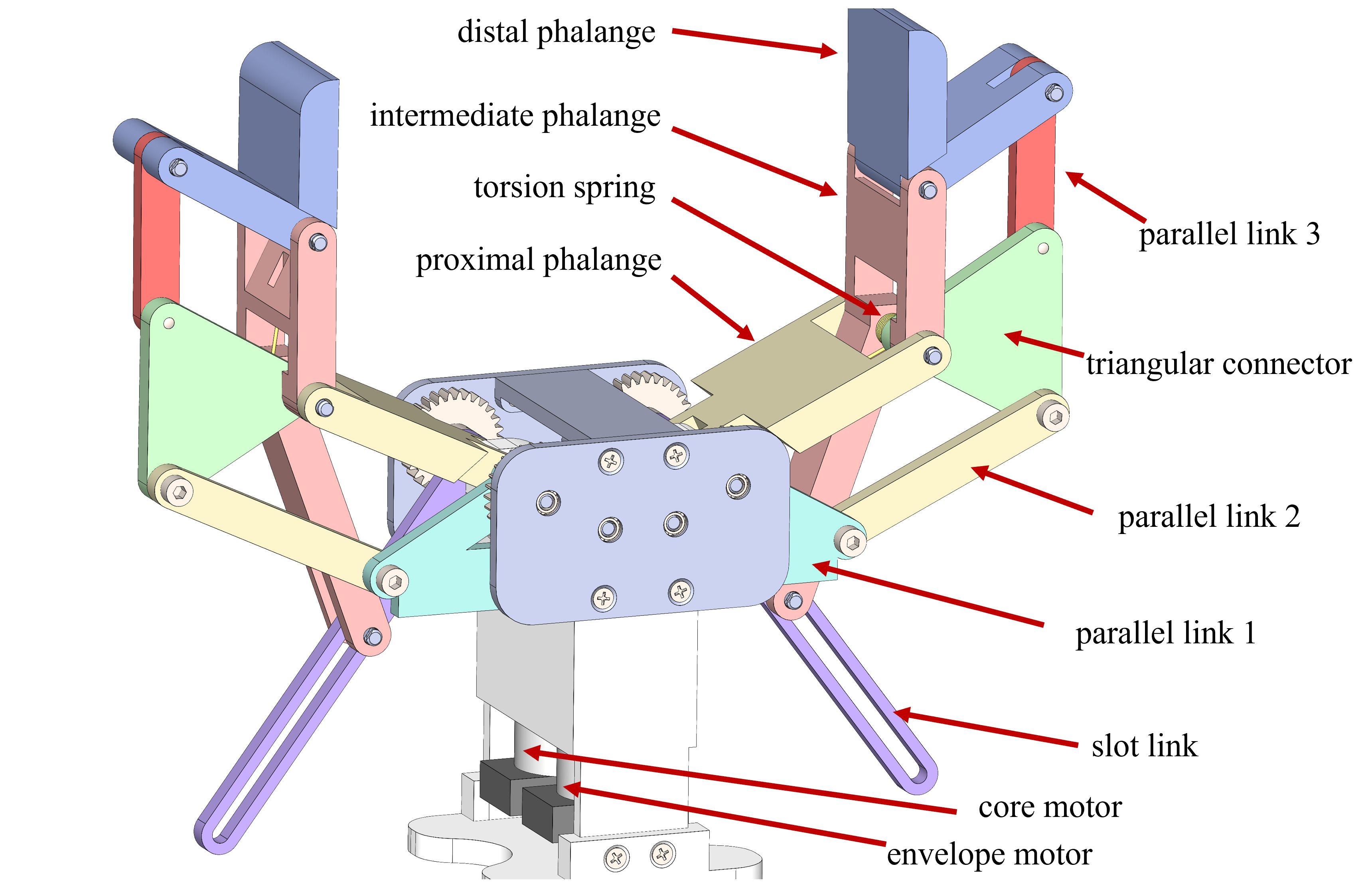}
  \caption{SCAL-R side view of the whole hand}
  \label{fig:scalr-side}
\end{figure}

Figure~\ref{fig:scalr-side} shows the side view with the main subassemblies. 
The finger follows the SCAL-based architecture with two parallelogram frames and a rigid \emph{triangular connector} linking them. 
A torsion spring mounted on the slot slider biases the mechanism toward the near end of the slot so that, in free space, the assembly acts as a rigid body during approach. 
We annotate the three phalanges (proximal, intermediate, distal), the parallelogram members in the lower and upper frames, and the connector that kinematically links them: 
\textbf{parallel link~1} denotes the lower short member (AE), while \textbf{parallel link~2} denotes its counterpart in each frame (EF in the lower frame and GH in the upper frame). 
The distal phalange is rigidly attached to the upper short member to maintain vertical orientation.

\begin{figure}[t]
  \centering
  \includegraphics[width=0.80\linewidth]{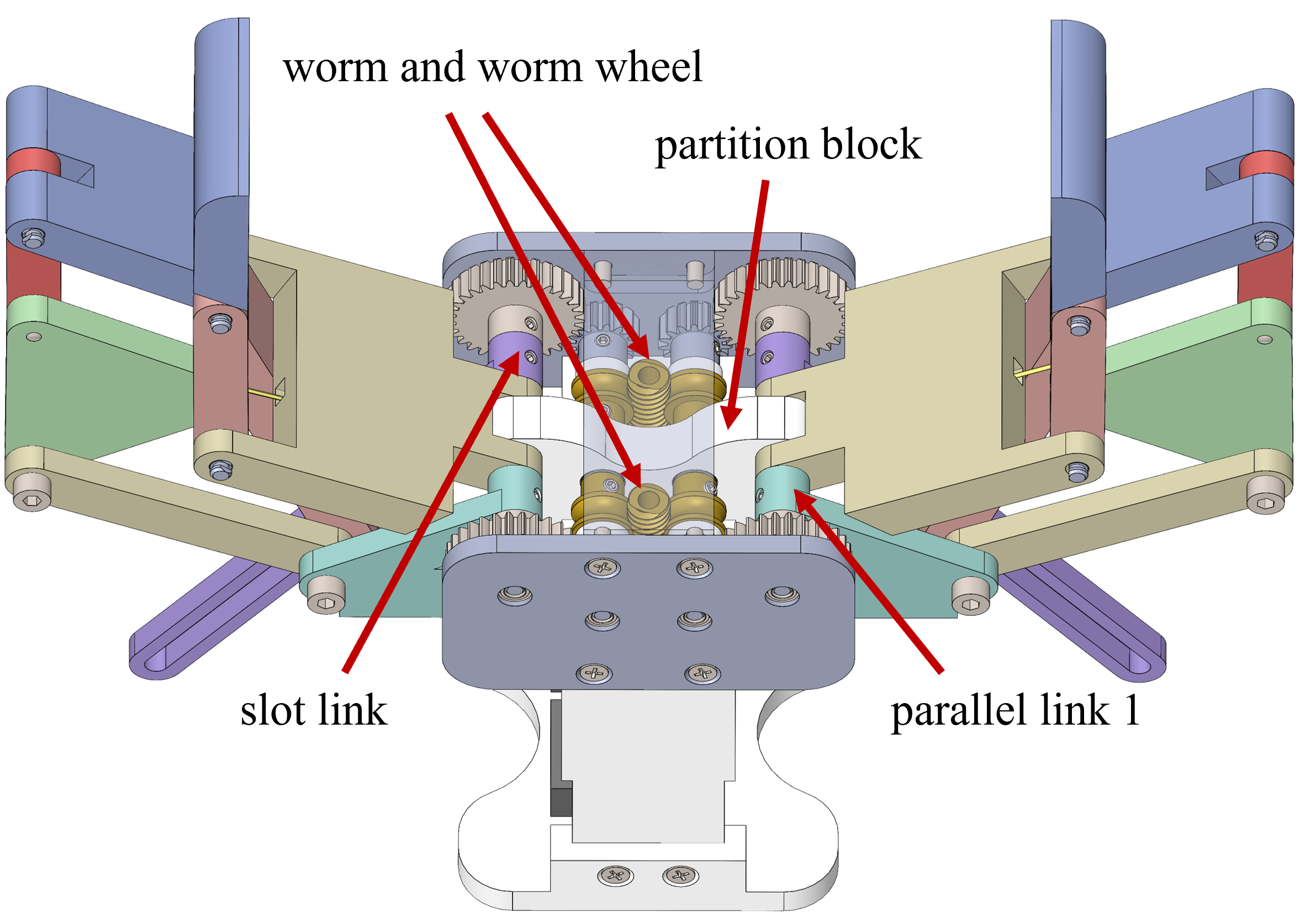}
  \caption{SCAL-R top view of the drive layout}
  \label{fig:scalr-top}
\end{figure}

Figure~\ref{fig:scalr-top} presents a top view of the drive layout. 
Two compact worm-gear stages are arranged symmetrically on the base plate. 
The \emph{core motor} drives a gear train that engages the \emph{slot link}, producing the main closing motion of the finger. 
The \emph{envelope motor} drives another worm-gear stage that turns the \emph{parallel link~1}; this motion is relayed through the triangular connector and the upper parallelogram to the distal phalange, enabling post-contact envelope and orientation control. 
The two drive trains are mechanically isolated by a central partition block, so that each motor drives its own side’s linkage symmetrically without mechanical interference, ensuring independent operation of the core and envelope functions.

\subsection{SCAL-L: Mechanical Design}

\begin{figure}[t]
  \centering
  \includegraphics[width=0.95\linewidth]{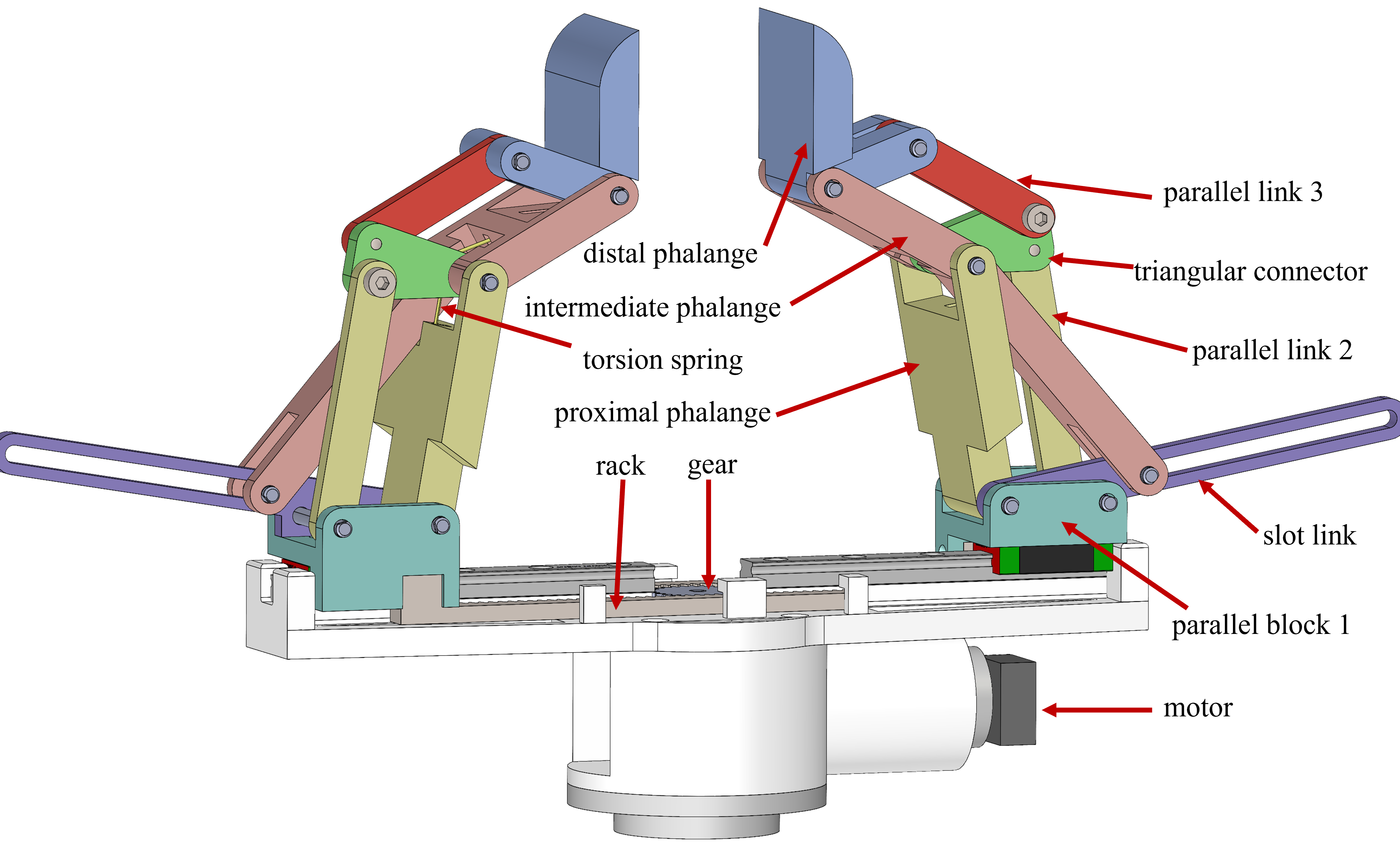}
  \caption{SCAL-L side view of the whole hand}
  \label{fig:scall-side}
\end{figure}

Figure~\ref{fig:scall-side} shows the side view of the SCAL-L prototype. 
This design shares the same finger architecture as SCAL-R: each finger consists of a slot link with a torsion-biased slider, a triangular connector bridging the lower and upper parallelogram frames, and three phalange segments (proximal, intermediate, distal). 
The parallelogram members (\textbf{parallel link~2}) maintain the distal phalange vertical, while the triangular connector ensures synchronous rotation of the upper and lower frames.

Unlike SCAL-R, SCAL-L integrates both fingers on a common horizontal carriage driven by a single motor. 
The motor rotates a gear that meshes with a rack mounted beneath the carriage. 
The rack is rigidly fastened to a large \textbf{parallel block~1} on each finger (replacing the previous parallel link~1), so that the block—and thus the entire finger—translates horizontally along linear guide rails. 
This linear-drive layout provides wide-aperture opening and synchronized closure while retaining the same surface-adaptive slot mechanism for contact compliance.

\section{ANALYSIS}

We analyze fingertip normal forces for the SCAL finger in two grasping modes: (i) linear parallel pinching (distal-only contact) and (ii) enveloping (simultaneous middle–distal contact with spring preload). The analysis is carried out at the \emph{finger level} and is independent of the actuation style. We assume quasi-static conditions and neglect contact friction; forces are obtained from moment balance and virtual work.

\subsection{Force Analysis in Linear Parallel Pinching}

\begin{figure}[t]
  \centering
  \includegraphics[width=0.4\linewidth]{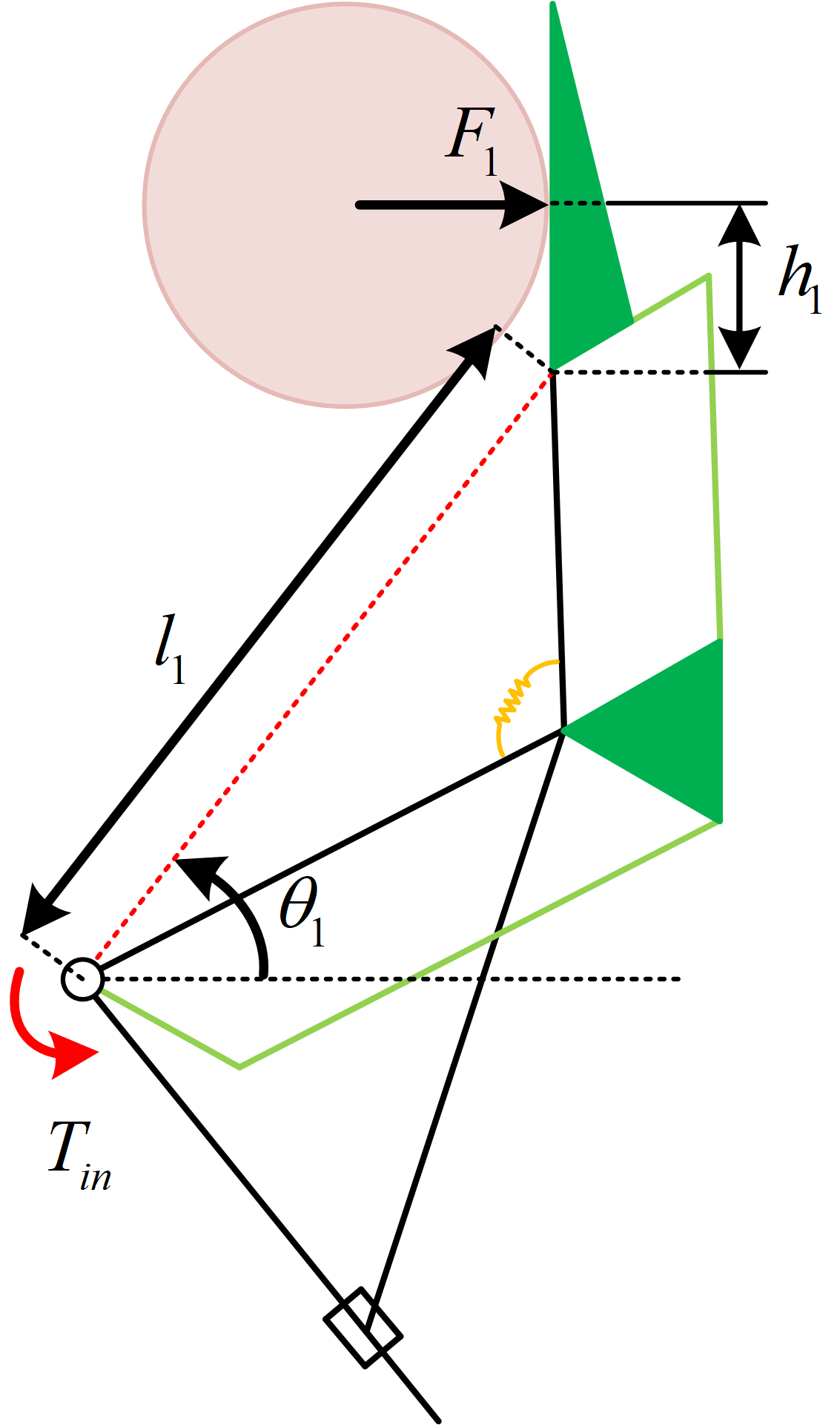}
  \caption{Force analysis of parallel pinch.}
  \label{fig:pinch_force}
\end{figure}

During parallel pinching the lower SCAL subchain (slot slider and inner joints) is effectively locked by the torsion spring, so the finger can be treated as a rigid linkage about the base. Figure~\ref{fig:pinch_force} shows the free-body geometry. Let $F_1$ be the fingertip normal force at the distal phalange, $h_1$ the perpendicular distance from the distal contact to the distal-joint axis (lever arm at the contact), $l_1$ the distance from the base to the distal joint along the rigidized link, and $\theta_1$ the link angle measured from the horizontal. Static moment balance about the base gives
\[
F_1\!\left(h_1 + l_1\sin\theta_1\right) \;=\; T_{\mathrm{in}},
\]
which yields the closed form
\[
\;F_1 \;=\; \dfrac{T_{\mathrm{in}}}{\,h_1 + l_1\sin\theta_1\,}\;
\]
This expression directly maps the equivalent input moment to the fingertip normal force and will be evaluated along the grasp path using the kinematic sequence $\theta_1(t)$.

\begin{figure}[t]
  \centering
  \includegraphics[width=0.85\linewidth]{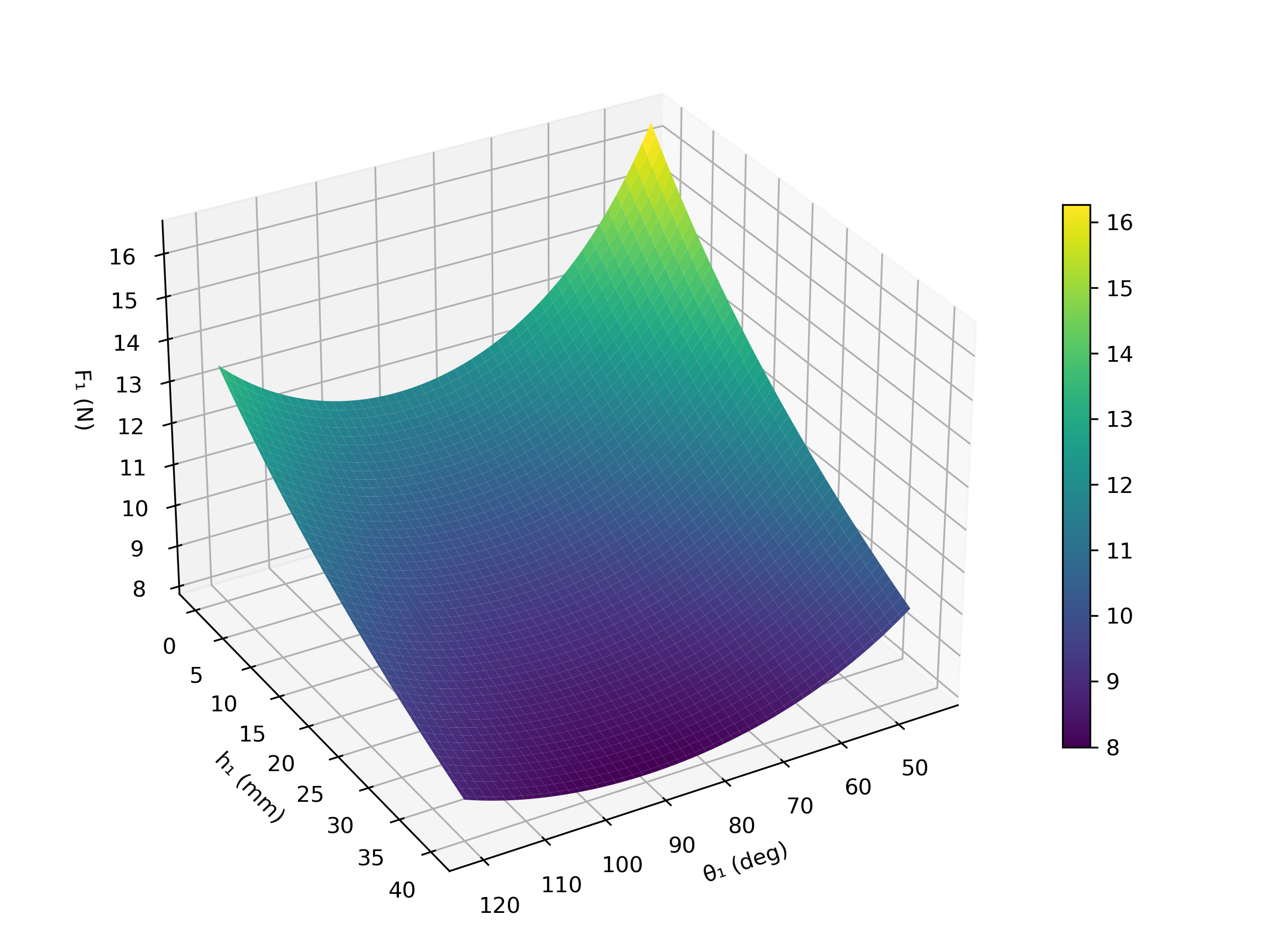}
  \caption{Fingertip force surface from $F_1=T_{\mathrm{in}}/(h_1+l_1\sin\theta_1)$,
  evaluated with $l_1=85.27$\,mm and $T_{\mathrm{in}}=1000$\,N·mm over
  $\theta_1\!\in[45^\circ,120^\circ]$, $h_1\!\in[0,40]$\,mm.}
  \label{fig:F1_surface}
\end{figure}

To illustrate this dependence, Fig.~\ref{fig:F1_surface} plots the predicted fingertip force over
$\theta_1\!\in[45^\circ,120^\circ]$ and $h_1\!\in[0,40]$\,mm for $l_1=85.27$\,mm and
$T_{\mathrm{in}}=1000$\,N·mm. The surface decreases with $h_1$ and exhibits a shallow minimum near
$\theta_1\!\approx\!90^\circ$, rising toward the span extremes; values scale linearly with
$T_{\mathrm{in}}$.

\subsection{Force Analysis in Enveloping}

In the enveloping mode, the \emph{proximal} and \emph{intermediate} phalanges contact the object simultaneously.  
We use two generalized coordinates: $\theta_2$ for the proximal joint and $\theta_3$ for the intermediate joint (counterclockwise positive).  
A torsional spring of stiffness $k_1$ biases the intermediate joint, yielding a resisting torque $T_s = k_1\theta_3$.  
The equivalent input moment at the base is denoted $T_{\mathrm{in}}$.

Let $G_1$ be the proximal-phalange contact (force $F_2$) and $G_2$ the intermediate-phalange contact (force $F_3$).  
Denote by $h_2$ the perpendicular distance from $G_1$ to the proximal-joint axis, by $h_3$ the perpendicular distance from $G_2$ to the intermediate-joint axis, and by $l_2$ the distance from the proximal to the intermediate joint.

\textit{Virtual displacements.}
Projected along each contact-force line, the virtual displacements are
\begin{align}
\delta s_1 &= h_2\,\delta\theta_2, \label{eq:env_ds1}\\
\delta s_2 &= l_2\cos(\theta_3-\theta_2)\,\delta\theta_2 \;+\; h_3\,\delta\theta_3. \label{eq:env_ds2}
\end{align}
The term $l_2\cos(\theta_3-\theta_2)$ arises because the velocity of $G_2$ induced by a small rotation $\delta\theta_2$ about the proximal joint is perpendicular to the proximal link with magnitude $l_2\,\delta\theta_2$; its projection onto the local contact normal at $G_2$ introduces the factor $\cos(\theta_3-\theta_2)$ (relative orientation between that velocity direction and the contact normal).

\textit{Virtual work and force mapping.}
Quasi-static balance (input power equals contact power) gives
\begin{align}
\begin{bmatrix} T_{\mathrm{in}} & -\,k_1\theta_3 \end{bmatrix}
\begin{bmatrix} \delta\theta_2 \\ \delta\theta_3 \end{bmatrix}
&=
\begin{bmatrix} F_2 & F_3 \end{bmatrix}
\underbrace{\begin{bmatrix}
h_2 & 0\\
l_2\cos(\theta_3-\theta_2) & h_3
\end{bmatrix}}_{J(\theta_2,\theta_3)}
\begin{bmatrix} \delta\theta_2 \\ \delta\theta_3 \end{bmatrix}. \label{eq:env_vw}
\end{align}
Therefore

\begin{align}
\begin{bmatrix} F_2 & F_3 \end{bmatrix}
&=
\begin{bmatrix} T_{\mathrm{in}} & -\,k_1\theta_3 \end{bmatrix}
\begin{bmatrix}
\displaystyle \frac{1}{h_2} &
\displaystyle 0 \\[6pt]
\displaystyle -\frac{l_2\cos(\theta_3-\theta_2)}{h_2h_3} &
\displaystyle \frac{1}{h_3}
\end{bmatrix}
\notag \\[6pt]
&\equiv
\begin{bmatrix} T_{\mathrm{in}} & -\,k_1\theta_3 \end{bmatrix}
J^{-1}(\theta_2,\theta_3).
\label{eq:env_inv}
\end{align}

Expanding \eqref{eq:env_inv} yields compact closed forms:
\begin{align}
\;F_2 \;=\; \frac{T_{\mathrm{in}}}{h_2} \;+\; \frac{k_1\theta_3\,l_2\cos(\theta_3-\theta_2)}{h_2h_3} \quad
\;F_3 \;=\; -\,\frac{k_1\theta_3}{h_3}\; \label{eq:env_closed}
\end{align}

\begin{figure}[t]
  \centering
  \includegraphics[width=0.60\linewidth]{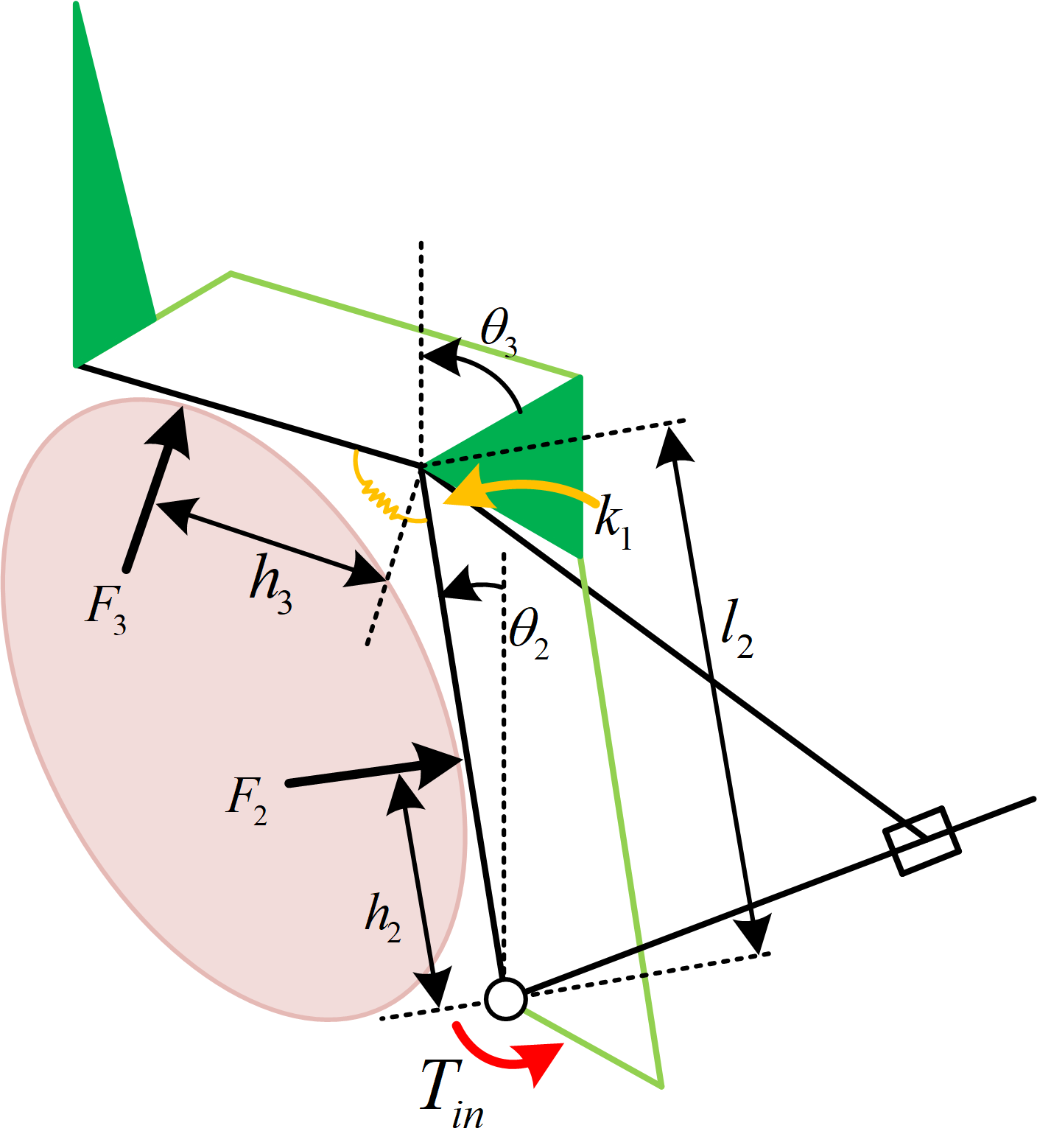}
  \caption{Force analysis of enveloping with proximal (force $F_2$) and intermediate (force $F_3$) contacts.}
  \label{fig:env_force}
\end{figure}

\begin{figure}[t]
  \centering
  \includegraphics[width=0.95\linewidth]{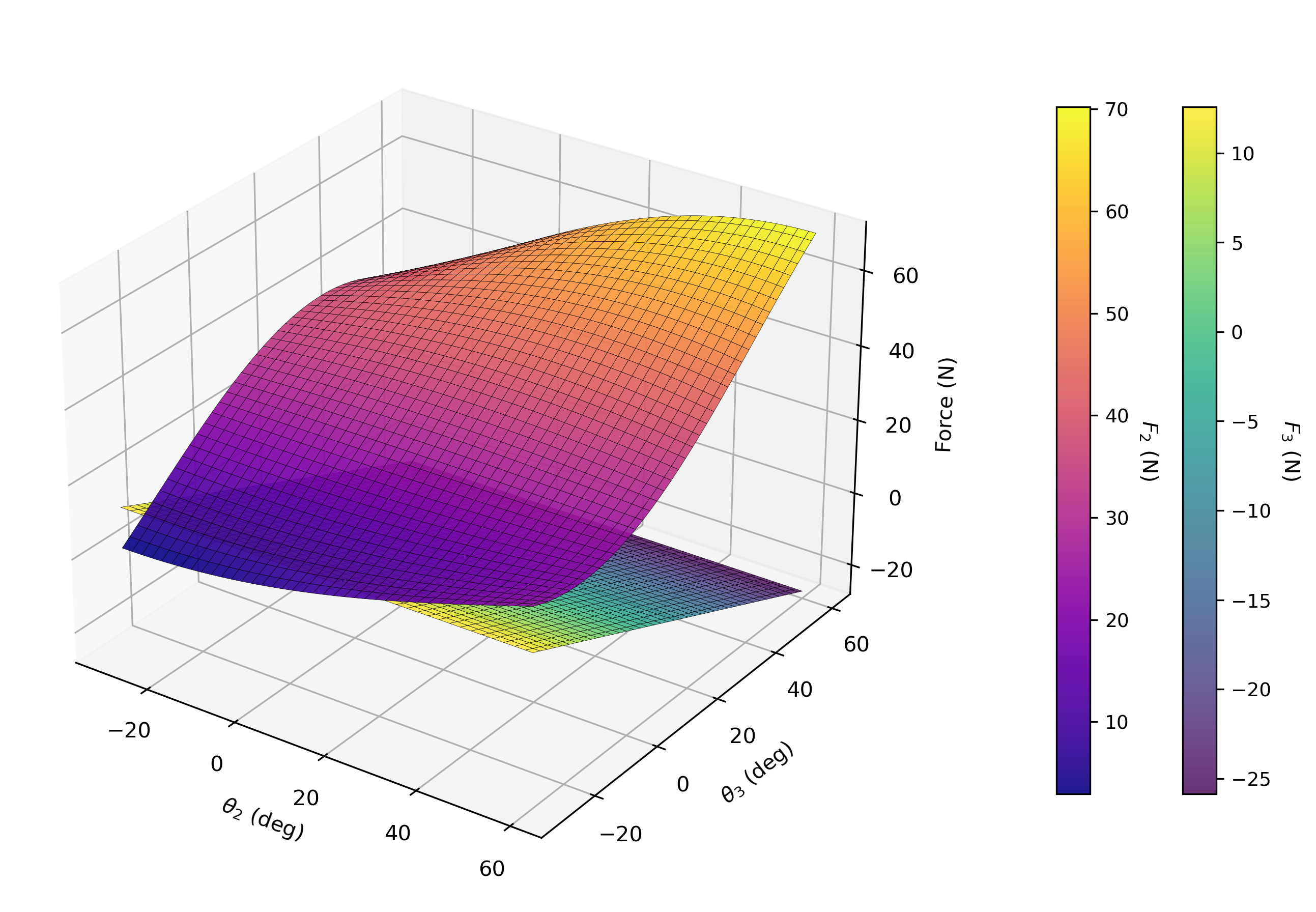}
  \caption{Enveloping contact forces versus joint angles using
  $T_{\mathrm{in}}=1000$\,N·mm, $k_1=500$\,N·mm/rad, $l_2=70$\,mm, $h_2=40$\,mm, $h_3=20$\,mm.}
  \label{fig:F23_surface}
\end{figure}

To visualize load sharing in enveloping, Fig.~\ref{fig:F23_surface} sweeps
$\theta_2\!\in[-30^\circ,60^\circ]$ and $\theta_3\!\in[-30^\circ,60^\circ]$
with $T_{\mathrm{in}}=1000$\,N·mm, $k_1=500$\,N·mm/rad, $l_2=70$\,mm, $h_2=40$\,mm, and $h_3=20$\,mm.
Consistent with \eqref{eq:env_closed}, the proximal-contact force $F_2$ contains a constant
baseline $T_{\mathrm{in}}/h_2$ (here $\approx25$\,N) plus a geometry–spring coupling term
that grows when $\theta_3\!\approx\!\theta_2$ (i.e., $\cos(\theta_3-\theta_2)\!\to\!1$) and for
positive $\theta_3$. The intermediate-contact force $F_3$ varies linearly with $\theta_3$
with slope $-k_1/h_3$ (here $\approx-25$\,N/rad), and its sign follows the local normal convention.

\section{EXPERIMENTS}

We validated the proposed designs with a full-scale, 3D-printed two-finger gripper.  
The experiments aimed to assess three key capabilities:  
(i) contact-preserving surface sliding and pinch-lifting,  
(ii) environment-adaptive alignment on sloped or curved supports, and  
(iii) grasping of bulky or wide objects via passive or active aperture modulation.  

Across both prototypes, over 50 grasp trials were conducted on diverse objects (thin parts, jars, tapes, boxes) with typical success rates above 90\%.  
Each sequence is shown in several key phases (contact, slide/align, lift/envelope) to highlight how contact constraints are exploited.  
We report the results separately for SCAL-R (rotational drive with active fingertip) and SCAL-L (linear drive with passive opening).

\subsection{SCAL-R (Rotational Drive with Active Fingertip)}

\begin{figure}[t]
  \centering
  \includegraphics[width=0.98\linewidth]{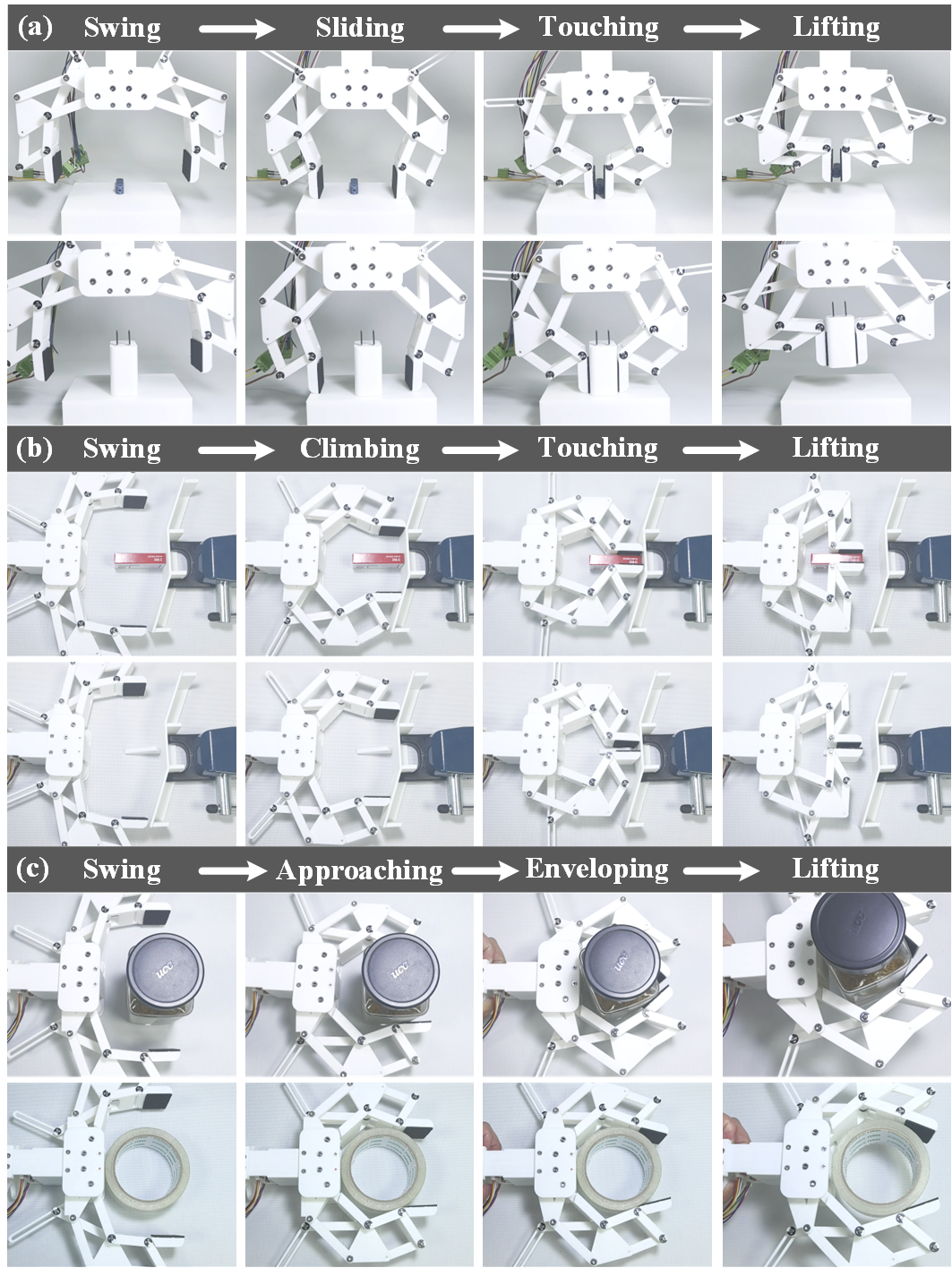}
  \caption{Grasping experiments with SCAL-R.  
  (a) Pinch-lifting on a flat surface.  
  (b) Ramp negotiation on a $20^\circ$ slope.  
  (c) Enveloping grasp on large objects.}
  \label{fig:r_exp}
\end{figure}

We evaluated the SCAL-R prototype in three representative tasks (Fig.~\ref{fig:r_exp}).  
Each sequence shows two object examples (rows) and four phases (columns).

\textbf{(a) Environment-assisted pinch-lifting on flat surfaces.}  
The fingertip first swings down to contact the tabletop, then slides while maintaining parallel pinching.  
Upon reaching the target, continued rotation bends the fingertip path upward, lifting the object from the surface.  
This demonstrates stable grasping of thin or low-profile items such as small parts of about $10\,\mathrm{mm}$ in height.

\textbf{(b) Ramp negotiation.}  
With the support inclined at $20^\circ$, one finger climbs the uphill side while the other descends the opposite slope.  
Both fingertips preserve contact as they converge and then transition upward to complete lifting, successfully grasping small box-like objects.

\textbf{(c) Enveloping large objects.}  
The finger first swings toward the object and establishes contact via the intermediate phalange; the distal phalange then continues forward, and the active fingertip folds inward to form an enveloping grasp.  
This enables secure grasping of bulky items such as a seasoning jar and a roll of transparent tape.

\subsection{SCAL-L (Linear Drive with Passive Opening)}

\begin{figure}[t]
  \centering
  \includegraphics[width=0.98\linewidth]{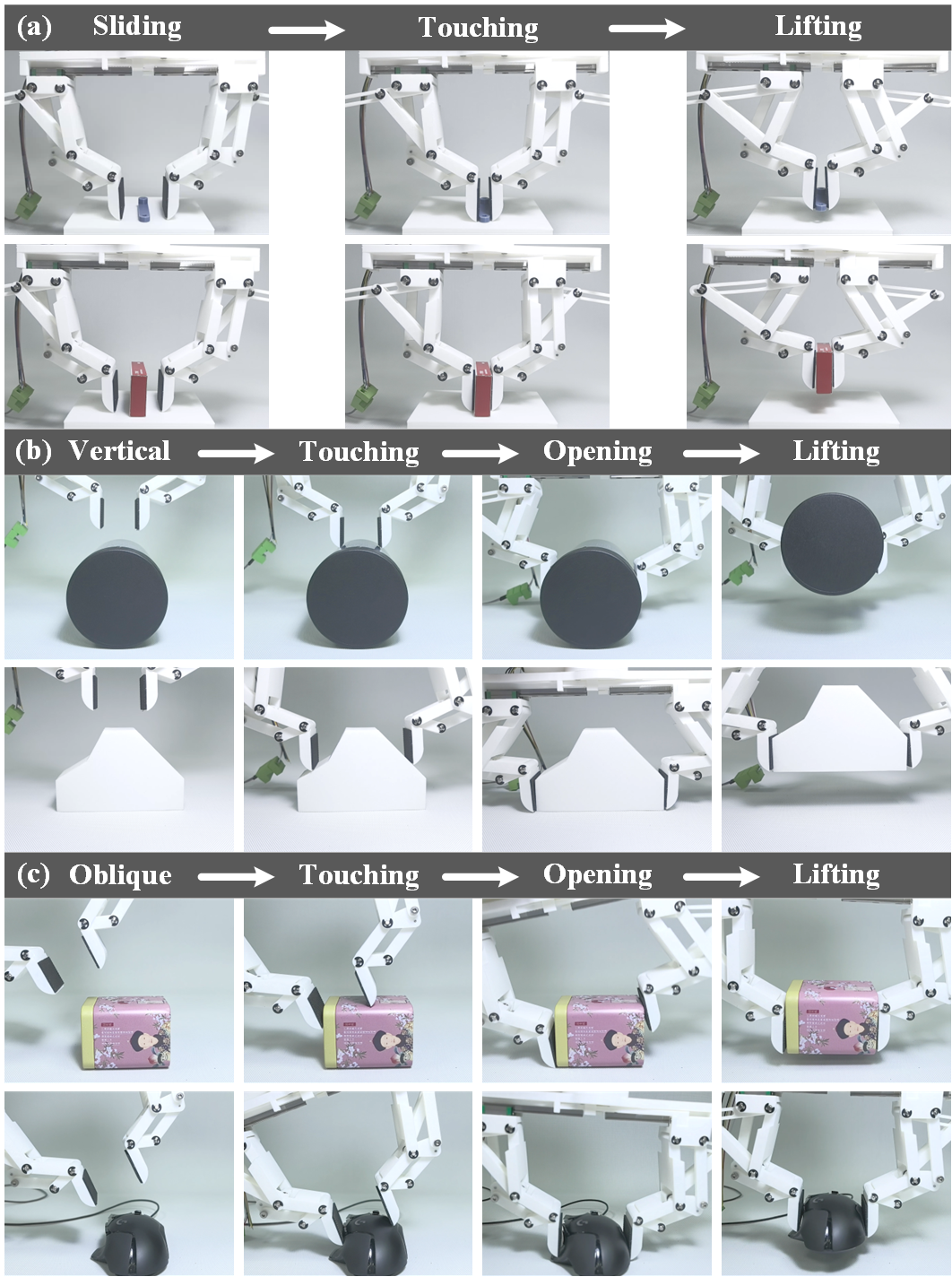}
  \caption{Grasping experiments with SCAL-L. 
  (a) Pinch-lifting by horizontal sweep. 
  (b) Vertical probing with passive opening. 
  (c) Oblique probing to trigger opening on flat faces.}
  \label{fig:l_exp}
\end{figure}

We evaluated the SCAL-L prototype in three representative behaviors (Fig.~\ref{fig:l_exp}). Each panel shows two object examples and a short sequence with three or four phases.

\textbf{(a) Pinch-lifting via horizontal sweep.}  
With translational drive, the finger sweeps across the tabletop, establishes contact, and slides while maintaining parallel pinching. As translation continues, the fingertip path turns upward and the object is lifted off the support.  
This demonstrates surface-assisted grasping of thin or low-profile parts without high-precision sensing.

\textbf{(b) Vertical probing with passive opening.}  
When probing straight down on curved or oblique surfaces, the contact reaction drives the slot slider outward, passively widening the aperture before the carriage completes closure and lift.  
This “vertical-first” strategy reliably captures round jars and chamfered blocks while preserving a simple control sequence.

\textbf{(c) Oblique probing for flat-faced objects.}  
If a flat face does not open under pure vertical probing, the finger approaches at a slight angle to engage an edge/corner or an inclined patch that triggers passive opening; once the span exceeds the object width, the horizontal drive closes to secure the grasp.  
This oblique-probe tactic extends applicability to flat tins and other weak-feature geometries.

In summary, both prototypes exploit contact to reduce perception and control burden while reliably transitioning from surface alignment to lifting. SCAL\texttt{-}R excels when an enveloping lock is desirable: the active fingertip secures bulky items after initial stabilization, at the cost of a slightly more complex drive. SCAL\texttt{-}L favors simplicity and large-aperture adaptation: translation plus passive opening enables probing-driven capture of wide or weak-feature objects without additional sensing. Typical failure cases observed across both systems involve extremely low-friction surfaces and compliance-limited edge contacts, which we mitigate by modest preload and approach speed tuning. Overall, the results indicate complementary operating regimes for the two actuation paradigms under environment-assisted grasping.

\section{Conclusions}
This work proposes an environment-assisted grasping strategy based on \emph{pinch-lifting}, implemented through a slot-constrained adaptive linkage (SCAL). Two prototypes are developed: SCAL-R uses rotational drive with an active fingertip, while SCAL-L employs linear drive with passive opening. Both designs demonstrate reliable surface alignment, lifting, and geometric adaptation, while reducing the need for sensing and control.

Future work will focus on quantitative benchmarking and deeper exploration of adaptive behaviors. In particular, we aim to characterize the grasping force profile under different geometries and approach strategies, and to optimize the linkage for improved robustness. These directions are expected to broaden the applicability of SCAL-based designs to more diverse, contact-rich manipulation tasks.

\addtolength{\textheight}{-12cm}   


\begin{thebibliography}{99}

    \bibitem{exploit2015} C. Eppner, R. Deimel, J. Álvarez-Ruiz \textit{et al.}, “Exploitation of environmental constraints in human and robotic grasping,” \textit{Int. J. Robot. Res.}, vol. 34, no. 7, pp. 1021-1038, 2015.
    \bibitem{underactuated2007} L. Birglen, T. Laliberté, and C. Gosselin, \textit{Underactuated Robotic Hands}. Berlin, Germany : Springer, 2007.
    \bibitem{AnalysisUnderactuated2011} R. Balasubramanian and A. M. Dollar, “Performance of serial underactuated mechanisms: Number of degrees of freedom and actuators,” in \textit{Proc. IEEE/RSJ Int. Conf. Intell. Robots Syst.}, 2011, pp. 1823-1829.
    \bibitem{robotiq} Demers A., Lefrançois L. A., Jobin S., \textit{et al.}, “Gripper Having a Two Degree of Freedom Underactuated Mechanical Finger for Encompassing and Pinch Grasping”. US Patent. US8973958B2. 2015-03-10.
    \bibitem{GR2} N. Rojas, R. R. Ma, and A. M. Dollar, “The GR2 gripper: An underactuated hand for open-loop in-hand planar manipulation,” \textit{IEEE Trans. Robot.}, vol. 32, no. 3, pp. 763-770, Jun. 2016.
    \bibitem{uGRIPP} A. Kobayashi, K. Yamaguchi, J. Kinugawa, S. Arai, Y. Hirata, and K. Kosuge, “Analysis of precision grip force for uGRIPP (underactuated gripper for power and precision grasp),” in \textit{Proc. IEEE/RSJ Int. Conf. Intell. Robots Syst.}, 2017, pp. 1937-1942.
    \bibitem{ParaGripper} H. Liu, L. Zhao, B. Siciliano, and F. Ficuciello, “Modeling, optimization, and experimentation of the ParaGripper for in-hand manipulation without parasitic rotation,” \textit{IEEE Robot. Autom. Lett.}, vol. 5, no. 2, pp. 3011-3018, Apr. 2020.
    \bibitem{PASA2016} Liang, D., Song, J., Zhang, W., \textit{et al}. “PASA Hand: A Novel Parallel and Self-Adaptive Underactuated Hand with Gear-Link Mechanisms”. \textit{Int. Conf. on Intelligent Robotics and Applications.} vol 9834., 2016.
    \bibitem{proximity2018} K. Sasaki, K. Koyama, A. Ming, M. Shimojo, R. Plateaux, and J. Choley, “Robotic grasping using proximity sensors for detecting both target object and support surface,” in \textit{Proc. IEEE/RSJ Int. Conf. Intell. Robots Syst.}, 2018, pp. 2925-2932.
    \bibitem{tool2025} M. Y. Aoyama, S. Vijayakumar and T. Narita, “Few-Shot Transfer of Tool-Use Skills Using Human Demonstrations With Proximity and Tactile Sensing,” \textit{IEEE Robot. Autom. Lett.}, vol. 10, no. 9, pp. 9024-9031, Sept. 2025.
    \bibitem{Velo2014} M. Ciocarlie, “The Velo gripper: A versatile single-actuator design for enveloping, parallel and fingertip grasps,” \textit{Int. J. Robot. Res.}, vol. 33, no. 5, pp. 753-767, 2014.
    \bibitem{BLT2020} Y. J. Kim, H. Song, and C. Y. Maeng, “BLT gripper: An adaptive gripper with active transition capability between precise pinch and compliant grasp,” \textit{IEEE Robot. Autom. Lett.}, vol. 5, no. 4, pp. 5518-5525, Oct. 2020.
    \bibitem{analysis2021} D. Yoon and Y. Choi, “Analysis of fingertip force vector for pinch-lifting gripper with robust adaptation to environments,” \textit{IEEE Trans. Robot.}, vol. 37, no. 4, pp. 1127-1143, Aug. 2021.
    \bibitem{three-finger2021} L. Kang, Y. Yang, J. Yang, and B.-J. Yi, “A three-fingered adaptive gripper with multiple grasping modes,” in \textit{Proc. 2021 IEEE/RSJ Int. Conf. Intell. Robots Syst.}, 2021, pp. 6097-6103.
    \bibitem{omega2022} D. Yoon and K. Kim, “Fully passive robotic finger for human-inspired adaptive grasping in environmental constraints,” \textit{IEEE/ASME Trans. Mechatron.}, vol. 27, no. 5, pp. 3841-3852, Oct. 2022.


\end{thebibliography}
\end{document}